\documentclass[11pt,a4paper]{article}
\usepackage[hyperref]{acl2018}
\usepackage{times}
\usepackage{latexsym}
\usepackage{graphicx}
\usepackage{amsmath}
\usepackage[hang, flushmargin]{footmisc}

\usepackage{url}

\aclfinalcopy % Uncomment this line for the final submission
\newenvironment{itemizesquish}{\begin{list}{\labelitemi}{\setlength{\itemsep}{0em}\setlength{\labelwidth}{0.5em}\setlength{\leftmargin}{\labelwidth}\addtolength{\leftmargin}{\labelsep}}}{\end{list}}

\makeatletter
\newcommand\footnoteref[1]{\protected@xdef\@thefnmark{\ref{#1}}\@footnotemark}
\makeatother

\title{Ontology Alignment in the Biomedical Domain \\
Using Entity Definitions and Context}

\author{
\parbox{\linewidth}{\centering
Lucy Lu Wang$^\dagger$, Chandra Bhagavatula, Mark Neumann, \\ Kyle Lo, Chris Wilhelm, and Waleed Ammar} \\
\\ Allen Institute for Artificial Intelligence
\\ $^\dagger$Department of Biomedical Informatics and Medical Education, University of Washington 
\\ Seattle, Washington, USA
\\ \texttt{lucylw@uw.edu}
}

\date{}

\begin{document}
\maketitle

\begin{abstract}
  Ontology alignment is the task of identifying semantically equivalent entities from two given ontologies. Different ontologies have different representations of the same entity, resulting in a need to de-duplicate entities when merging ontologies. 
  We propose a method for enriching entities in an ontology with external definition and context information, and use this additional information 
%   when computing entity embeddings 
  for ontology alignment.
  We develop a neural architecture capable of encoding the additional information when available, and show that the addition of external data results in an F1-score of 0.69 on the Ontology Alignment Evaluation Initiative (OAEI) largebio SNOMED-NCI subtask, comparable with the entity-level matchers in a SOTA system.
%\cscomment{Can we compare to state of the art here? Or say that the logistic regression baseline is a good approximation of sota ?} \klcomment{confused by this sentence. are you saying a neural network model w/out enrichment is X better than a logistic regression baseline, and the neural network model w/ enrichment is 4\% better on top of the unenriched model?}
\end{abstract}

% \section{Credits}

\section{Introduction}

% Ontologies are important for NLP tasks.
Ontologies are used to ground lexical items in various NLP tasks 
% in many NLP tasks to ground lexical items into entities, 
including entity linking, 
%\cite[e.g.,][]{zheng_linking_2010,rao_linking_2013}, 
question answering, 
%\cite[e.g.,][]{athenikos_survey_2010,bordes_qa_2014}, 
semantic parsing
%\cite[e.g.,][]{berant_parsing_2013,yih_parsing_2014} 
and information retrieval.\footnote{Ontological resources include ontologies, knowledgebases, terminologies, and controlled vocabularies. In the rest of this paper, we refer to all of these with the term `ontology' for consistency.}
%\cite[e.g.,][]{vallet_retrieval_2005,xiong_retrieval_2017}.
% Plenty of ontologies are available in biomedicine.
In biomedicine, an abundance of ontologies (e.g., MeSH, Gene Ontology) has been developed for different purposes. Each ontology describes a large number of concepts in healthcare, public health or biology, enabling the use of ontology-based NLP methods in biomedical applications.
% The ontologies have significant overlap.
However, since these ontologies are typically curated independently by different groups, many important concepts are represented inconsistently across ontologies (e.g., ``Myoclonic Epilepsies, Progressive'' in MeSH is a broader concept that includes ``Dentatorubral-pallidoluysian atrophy'' from OMIM).

% but it isn't clear how to consolidate multiple relevant ontologies.
This poses a challenge for bioNLP applications where multiple ontologies are needed for grounding, but each concept must be represented by only one entity. 
For instance, in \url{www.semanticscholar.org}, scientific publications related to carpal tunnel syndrome are linked to one of multiple entities derived from UMLS terminologies representing the same concept,\footnote{See \url{https://www.semanticscholar.org/topic/Carpal-tunnel-syndrome/248228} and \url{https://www.semanticscholar.org/topic/Carpal-Tunnel-Syndrome/3076}} making it hard to find all relevant papers on this topic.
% ontology matching.
To address this challenge, we need to automatically map semantically equivalent entities from one ontology to another. This task is referred to as ontology alignment or ontology matching.

\begin{figure*}[!htpb]
\begin{center}
\includegraphics[width=0.8\textwidth,keepaspectratio]{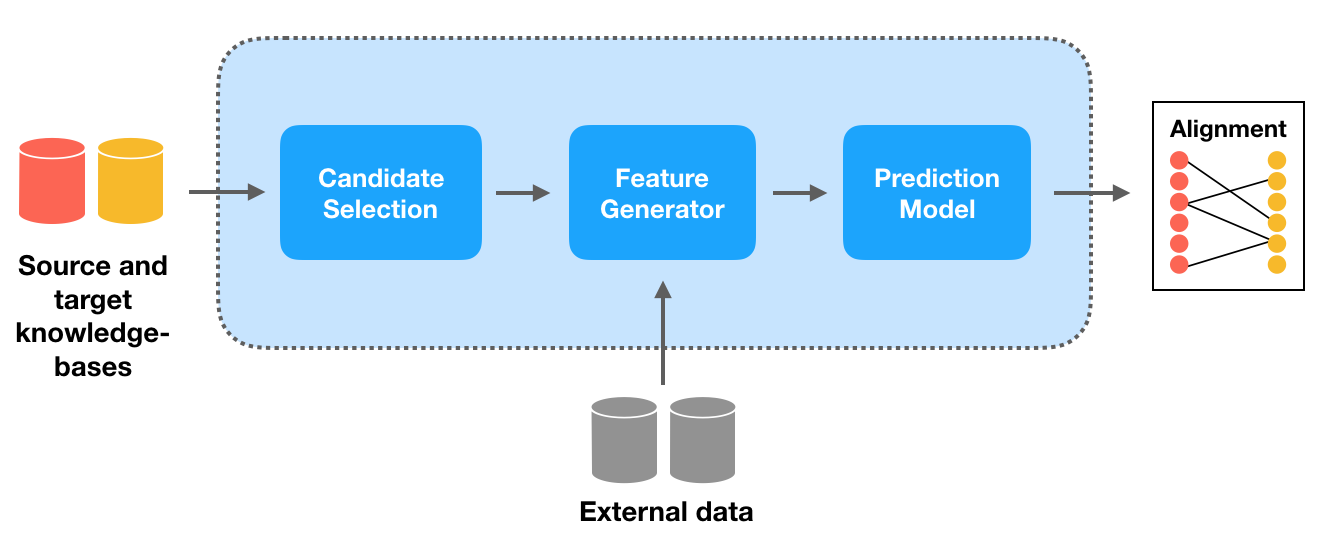}
\caption{\label{fig:ontoemma} OntoEmma consists of three modules: a) candidate selection (see \S\ref{sec:features} for details), b) feature generation (see \S\ref{sec:features} for details), and c) prediction (see \S\ref{sec:prediction} for deatils). OntoEmma accepts two ontologies (a source and a target) as inputs, and outputs a list of alignments between their entities. When using a neural network, the feature generation and prediction model are combined together in the network.}
% The candidate selection module generates an ordered list of candidate entity pairs from the two ontologies, from which we take the top $K$ candidates for each entity from the source ontology. The feature generation module computes a set of features for each candidate pair, and incorporates external data such as textual contexts or definitions of the entities. Finally, the prediction module processes each candidate pair and outputs a similarity score between 0.0 and 1.0.
\vspace{-0.5ex}
\end{center}
\end{figure*}

% Brief description of previous efforts.
Several methods have been applied to ontology alignment, including rule-based and statistical matchers.
Existing matchers rely on entity features such as names, synonyms, as well as relationships to other entities \cite{shvaiko_ontology_2013,otero-cerdeira_ontology_2015}. 
% Limitations of previous work, which motivate our work.
However, it is unclear how to leverage the natural language text associated with entities to improve predictions.
% Our contributions.
We address this limitation by incorporating two types of natural language information (definitions and textual contexts) in a supervised learning framework for ontology alignment. 
Since the definition and textual contexts of an entity often provide complementary information about the entity's meaning, we hypothesize that incorporating them will improve model predictions.
We also discuss how to automatically derive labeled data for training the model by leveraging existing resources.
In particular, we make the following contributions:
\begin{itemizesquish}
\item We propose a novel neural architecture for ontology alignment and show how to effectively integrate natural language inputs such as definitions and contexts in this architecture (see \S\ref{sec:model} for details).\footnote{\label{github}Implementation and data available at \url{https://www.github.com/allenai/ontoemma/}}
%\cscomment{@Lucy do you have a preference for having two separate links ? I'd suggest just pointing to github and have downloads link there. There's a PR for adding the downloads link to the readme.}
%\llwcomment{Sounds good to me}
\item We use the UMLS Metathesaurus to extract large amounts of labeled data for supervised training of ontology alignment models (see \S\ref{sec:umls_data}).
We release our data set to help future research in ontology alignment.\footnoteref{github}
%\footnote{Data is available at \url{http://llwang.net/data/ontoemma/}.}
\item We use external resources such as Wikipedia and scientific articles to find entity definitions and contexts (see \S\ref{sec:external_data} for details).
\end{itemizesquish}
%\klcomment{Unless the dataset is released, I don't really understand why the 2nd and 3rd bullets are considered ``contributions'' in the same way that the 1st bullet for novel neural architecture is a contribution.  }

\section{OntoEmma}\label{sec:model}
In this section, we describe OntoEmma, our proposed method for ontology matching, which consists of three stages: candidate selection, feature generation and prediction (see Fig. \ref{fig:ontoemma} for an overview).

%The following section describes the architecture of the ontology matcher, the architecture of the neural network model, the training data used to train the model, and the methods used for deriving mention contexts and enriching entity definitions. 

\subsection{Problem definition and notation}
We start by defining the ontology matching problem:
Given a source ontology $O^s$ and a target ontology $O^t$, each consisting of a set of entities, find all semantically equivalent entity pairs, i.e., $\{ (\mathbf{e}^s, \mathbf{e}^t) \in O^s \times O^t : \mathbf{e}^s \equiv \mathbf{e}^t\}$, where $\equiv$ indicates semantic equivalence. %\cscomment{nit: we might want to say $O^s$ is the set of entities from the source ontology instead of just calling it an ontology?}
%In addition to the unaligned ontologies $O^s$ and $O^t$, we are also given labeled pairs of semantically equivalent entity pairs from other ontologies for training, i.e., $\{ (\mathbf{e}^i, \mathbf{e}^j) \in O^i \times O^j: \mathbf{e}^i \equiv \mathbf{e}^j\}$ where $(O^i \cup O^j) \cap (O^s \cup O^t) = \Phi$. \klcomment{What is $\Phi$?} \klcomment{I think I understand this section better if you remove this ``In addition...''.  Seems only necessary to talk about training-data details when it gets to the section discussing training.}
For consistency, we preprocess entities from different ontologies to have the same set of attributes: a canonical name ($\mathbf{e}_{\textrm{name}}$), a list of aliases ($\mathbf{e}_{\textrm{aliases}}$), a textual definition ($\mathbf{e}_{\textrm{def}}$), and a list of usage contexts ($\mathbf{e}_{\textrm{contexts}}$).\footnote{Some of these attributes may be missing or have low coverage. See \S\ref{sec:external_data} for coverage details.}

%The goal of an ontology matcher is to take as input a pair of ontologies or terminologies, and output a list of positive matches between them. 
%Because the training data and the data we are interested in matching consist of a mixture of ontologies and terminologies (flat lists of terms without defined relationships or structure), we focus on matching entities between each pair of resources. 

\begin{figure*}[!ht]
\begin{center}
\includegraphics[width=\textwidth,keepaspectratio]{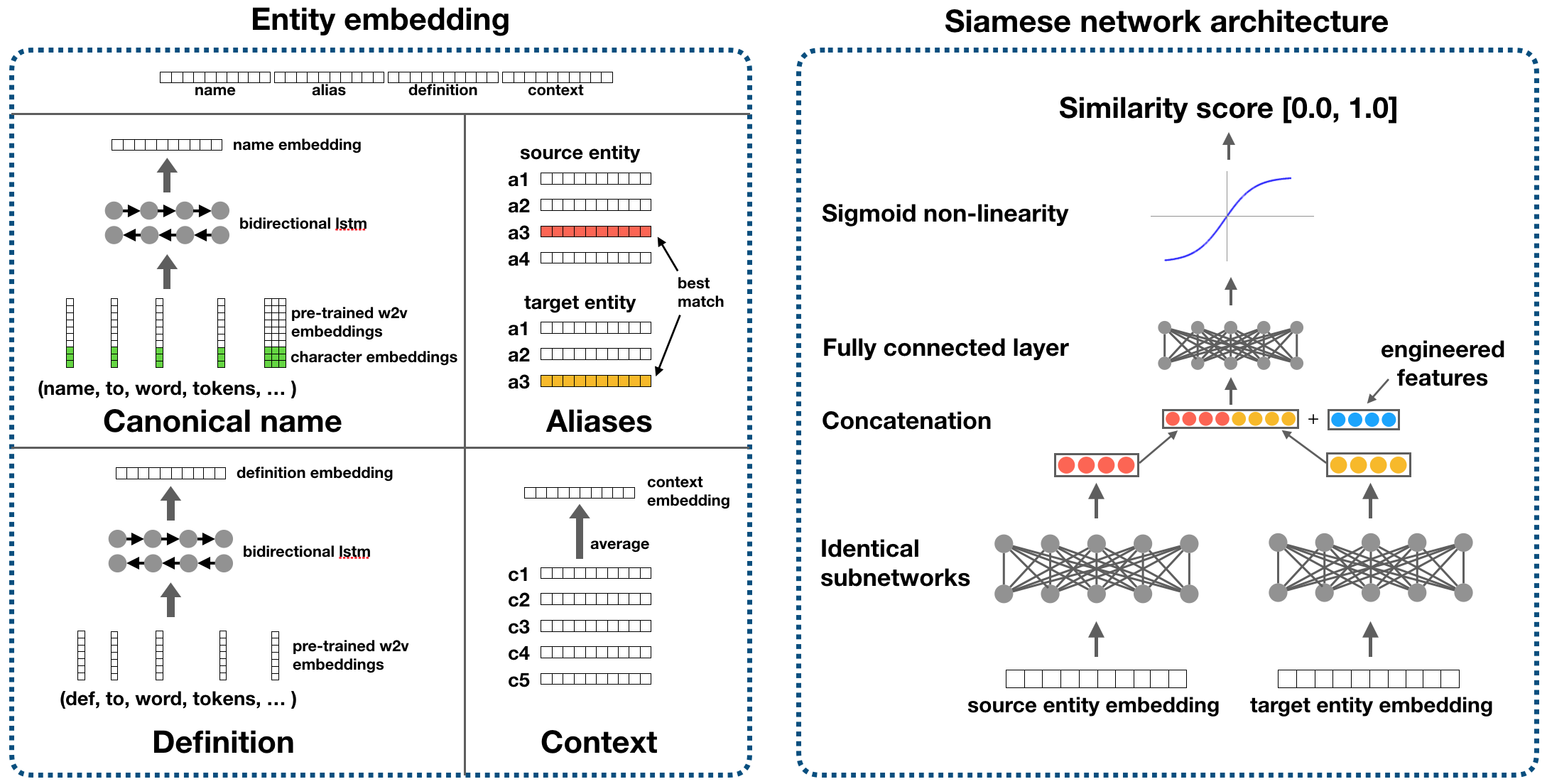}
\caption[width=0.8\textwidth]{\label{fig:nn} Siamese network architecture for computing entity embeddings for each source and target entity in a candidate entity pair.}
%Pretrained word embeddings are used at the word level, and character-based embeddings are concatenated to the pre-trained embedding vector for canonical name and aliases. The name, alias, definition, and context vectors are concatenated together to form the entity embedding. A similarity score is computed using a siamese network architecture (\emph{right}), which consists of two identical subnetworks for the source and target embeddings, concatenation of the outputs, a fully connected layer, and a sigmoid non-linearity. Engineered features were added during the concatenation step.}
%\wacomment{I think we need to revise the labels in this figure to be consistent with the standard way siamese networks are described. A siamese network consists of two identical subnetworks followed by a function which computes the similarity between the output of the two subnetworks. In our case, we define the similarity function to be concatenation, followed by two fully connected layers and a sigmoid non-linearity. }
%\cscomment{shall we call this a fully connected layer instead?}
%\lwcomment{I relabeled the figure; hopefully everything's clearer now.}
%\klcomment{The left one says ``Context from S2''.  Is that ok?}
\end{center}
\end{figure*}

\subsection{Candidate selection and feature generation}\label{sec:features}
%A framework for ontology matching called OntoEmma is developed in Python 3.6 (Figure \ref{fig:ontoemma}). We elect to use Python due to the availability of numerous popular natural language processing, machine learning, and deep learning libraries. Although there are fewer tools in Python for working with OWL and RDF files, the increased prototyping speed and deep learning support available in Python makes up for this weakness.\wacomment{Implementation details such as choice of programming language or libraries used can distract the reader from the main contributions. I've moved this paragraph to the experiment section.}

Many ontologies are large, which makes it computationally expensive to consider all possible pairs of source and target entities for alignment.
For example, the number of all possible entity pairs in our training ontologies is on the order of $10^{11}$.
In order to reduce the number of candidates, we use an inexpensive low-precision, high-recall candidate selection method using the inverse document frequency (\emph{idf}) of word tokens appearing in entity names and definitions.
For each source entity, we first retrieve all target entities that share a token with the source entity. Given the set of shared word tokens $\mathbf{w_{s+t}}$ between a source and target entity, we sum the \emph{idf} of each token over the set, yielding $idf_{total} = \sum_{i \epsilon \mathbf{w_{s+t}}} idf(i)$. Tokens with higher \emph{idf} values appear less frequently overall in the ontology and presumably contribute more to the meaning of a specific entity. We compute the \emph{idf} sum for each target entity and output the $K=50$ target entities with the highest value for each source entity, yielding $|O^s| \times K$ candidate pairs.

For each candidate pair ($\mathbf{e}^s, \mathbf{e}^t$), we precompute a set of 32 features commonly used in the ontology matching literature including the token Jaccard distance, stemmed token Jaccard distance, character n-gram Jaccard distance, root word equivalence, and other boolean and probability values over the entity name, aliases, and definition.\footnote{Even though neural models may obviate the need for feature engineering, feeding highly discriminative features into the neural model improves the inductive bias of the model and reduces the amount of labeled data needed for training.}

\subsection{Prediction}\label{sec:prediction}
Given a candidate pair ($\mathbf{e}^s,\mathbf{e}^t$) and the precomputed features $\mathbf{f}(\mathbf{e}^s,\mathbf{e}^t)$, we train a model to predict the probability that the two entities are semantically equivalent.
%The decision module, in the case of logistic regression and SVM, takes in the set of features for the candidate pair of entities, and produces a match probability score between 0.0 and 1.0, 1.0 being the highest likelihood of match. \wacomment{I removed this so that we can focus on the neural architecture in this part. However, please note that SVM is not a probabilistic model and therefore it doesn't output a probability.}
%For neural network models, both the feature generation and similarity prediction modules are replaced by a single neural network model. For each candidate pair, vectorized representations are computed for the source and target entities, and a similarity score is computed by the neural network.\wacomment{But we also use the precomputed features in some of the NN experiments, right?}
Figure \ref{fig:nn} illustrates the architecture of our neural model for estimating this probability which resembles a siamese network \cite{bromley_siamese_1993}.
At a high level, we first encode each of the source and target entities, then concatenate their representations and feed it into a multi-layer perceptron ending with a sigmoid function for estimating the probability of a match. 
Next, we describe this architecture in more detail.

\paragraph{Entity embedding.}
As shown in Fig. \ref{fig:nn} (\emph{left}), we encode the attributes of each entity as follows:
\begin{itemizesquish}
\item A canonical name $\mathbf{e}_\textrm{name}$ is a sequence of tokens, each encoded using pretrained \texttt{word2vec} embeddings concatenated with a character-level convolutional neural network (CNN). The token vectors feed into a bi-directional long short-term memory network (LSTM) and the hidden layers at both ends of the bi-directional LSTM are concatenated and used as the name vector $\mathbf{v}_\textrm{name}$.
\item Each alias in $\mathbf{e}_\textrm{aliases}$ is independently embedded using the same encoder used for canonical names (with shared parameters), yielding a set of alias vectors $\mathbf{v}_{\textrm{alias}-i}$ for $i=1, \ldots, |\mathbf{e}_\textrm{aliases}|$.
\item An entity definition $\mathbf{e}_\textrm{def}$ is a sequence of tokens, each encoded using pretrained embeddings then fed into a bi-directional LSTM. The definition vector $\mathbf{v}_\textrm{def}$ is the concatenation of the final hidden states in the forward and backward LSTMs.
\item Each context in $\mathbf{e}_\textrm{contexts}$ is independently embedded using the same encoder used for definitions (with shared parameters), then averaged yielding the context vector $\mathbf{v}_{\textrm{contexts}}$.
%\footnote{We discuss the coverage of definitions and contexts in \S\ref{sec:external_data}.} \wacomment{redundant info, mentioned before. removed it to fit within the page limit.}
\end{itemizesquish}

%\footnote{\lwcomment{@Chandra: I'm not sure about the details of the word embeddings. Were they GloVe embeddings trained over PubMed? PubMed Central?}\cscomment{Did you use this: http://bio.nlplab.org/ or something that I provided? If it was something I provided, it was Word2vec on S2 papers.}} 

The name, alias, definition, and context vectors are appended together to create the entity embedding, e.g., the source entity embedding $\mathbf{e}^s$ is:
$\mathbf{v}^s = [\mathbf{v}^s_\textrm{name};  
 \mathbf{v}^s_{\textrm{alias}-i^*};
 \mathbf{v}^s_\textrm{def};
 \mathbf{v}^s_\textrm{contexts}].
$
In order to find representative aliases for a given pair of entities, we pick the source and target aliases with the smallest Euclidean distance, i.e., $i^*, j^* = \arg\min_{i,j} \Vert \mathbf{v}^s_{\textrm{alias}-i} - \mathbf{v}^t_{\textrm{alias}-j} \Vert_2 $
%highest cosine similarity, i.e., $i^*, j^* = \arg\max_{i,j} \textrm{cos-sim}(\mathbf{v}^s_{\textrm{alias}-i}, \mathbf{v}^t_{\textrm{alias}-j})$. 
% \wacomment{Lucy, could you confirm whether we use cosine similarity here?} 
% \lwcomment{Actually, I used Euclidean distance. I replaced the equation above.}

\paragraph{Siamese network.}
After the source and target entity embeddings are computed, they are fed into two subnetworks with shared parameters followed by a parameterized function for estimating similarity.
Each subnetwork is a two layer feedforward network with ReLU non-linearities and dropout \cite{srivastava_dropout_2014}.
The outputs of the two subnetworks are then concatenated together with the engineered features and fed into another feedforward network with a ReLU layer followed by a sigmoid output layer.
We train the model to minimize the binary cross entropy loss for gold labels. 

To summarize, the network estimates the probability of equivalence between $\mathbf{e}^s$ and $\mathbf{e}^t$ as follows:
\begin{align}
\mathbf{h}^s &= \textsc{ReLU}(\textsc{ReLU}(\mathbf{v}^s; \theta_1); \theta_2) \nonumber \\
\mathbf{h}^t &= \textsc{ReLU}(\textsc{ReLU}(\mathbf{v}^t; \theta_1); \theta_2) \nonumber \\
P(\mathbf{e}^s \equiv \mathbf{e}^t) &= \textsc{Sigmoid}(\textsc{ReLU}([\mathbf{h}^s;\mathbf{h}^t]; \theta_3); \theta_4) \nonumber
\end{align}
%During prediction, only the similarity score is returned, which is then used by the OntoEmma framework to make a classification decision.

\section{Deriving and enriching labeled data} 
In this section, we discuss how to derive a large amount of labeled data for training, and how to augment entity attributes with definitions and contexts from external resources.

\subsection{Deriving training data from UMLS}\label{sec:umls_data}

The Unified Medical Language System (UMLS) Metathesaurus, which integrates more than 150 source ontologies, illustrates the breadth of coverage of biomedical ontologies \cite{bodenreider_unified_2004}. 
Also exemplified by the UMLS Metathesaurus is the high degree of overlap between the content of some of these ontological resources, whose terms have been (semi-)manually aligned. 
Significant time and effort has gone into cross-referencing semantically equivalent entities across the ontologies, and new terms and alignments continue to be added as the field develops. 
These manual alignments are high quality, but considered to be incomplete \cite{morrey_resolution_2011,mougin_auditing_2014}.

To enable supervised learning for our models, training data was derived from the UMLS Metathesaurus. By exposing our models to labeled data from the diverse subdomains covered in the UMLS Metathesaurus, we hope to learn a variety of patterns indicating equivalence between a pair of entities which can generalize to new ontologies not included in the training data.

We identified the following set of ontologies within UMLS to use as the source of our labeled data, such that they cover a variety of domains without overlapping with the test ontologies used for evaluation in the OAEI:
Current Procedural Terminology (CPT), Gene Ontology (GO), Hugo Nomenclature (HGNC), Human Phenotype Ontology (HPO), Medical Subject Headings (MeSH), Online Mendelian Inheritance in Man (OMIM), and RxNorm.

Our labeled data take the form ($\mathbf{e}^s,\mathbf{e}^t, l \in \{0, 1\}$), where $l=1$ indicates positive examples where $\mathbf{e}^s \equiv \mathbf{e}^t$. 
For each pair of ontologies, we first derive all the positive mappings from UMLS.
We retain the positive mappings for which there are no name equivalences.
Then, for each positive example ($\mathbf{e}^s,\mathbf{e}^t_+,1$), we sample negative mappings ($\mathbf{e}^s,\mathbf{e}^t_-,0$) from the other entities in the target ontology. 
One ``easy'' negative and one ``hard'' negative are selected for each positive alignment, where easy negatives consist of entities with little overlap in lexical features while hard negatives have high overlap. 
Easy negatives are obtained by randomly sampling entities from the target ontology, for each source entity.
Hard negatives are obtained using the same candidate selection method described in \S\ref{sec:model}. 
In both easy and hard examples, we exclude all target entities which appear in a positive example.\footnote{Although the negative examples we collect may be noisy due to the incompleteness of manual alignments in UMLS, this noise is also present in widely adopted evaluation of knowledge base completion problems and relation extraction with distant supervision \cite[e.g.,][]{li_kbc_2016,mintz_2009_distant}.}
%\wacomment{@Lucy: can we manually label a set of 100 hard negatives to get a rough estimate of how much noise we have due to incompleteness of alignments in UMLS?} \lwcomment{Yes, but I'm going to make this low priority.}\wacomment{got it, thanks!}

\begin{figure}[!htpb]
\begin{center}
\includegraphics[width=0.46\textwidth,keepaspectratio]{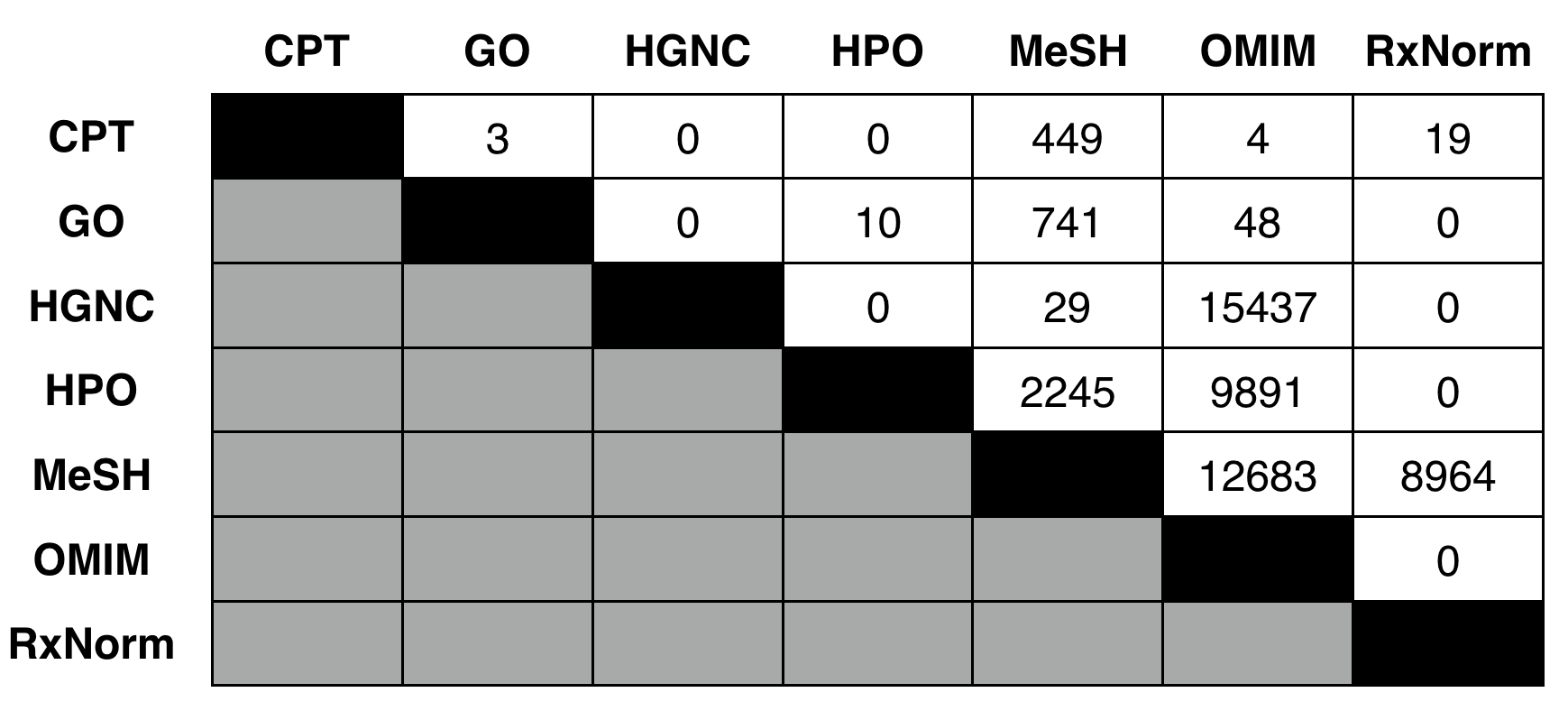}
\caption[width=0.8\textwidth]{\label{fig:posdist} Number of positive alignments extracted from each pair of ontologies from UMLS.}
\end{center}
\end{figure}

Over all seven ontologies, 50,523 positive alignments were extracted from UMLS. 
Figure \ref{fig:posdist} reports the number of positive alignments extracted from each ontology pair.
For these positives, 98,948 hard and easy negatives alignments were selected. 
These positive and negative labeled examples were pooled and randomly split into a 64\% training set, a 16\% development set, and a 20\% test set. 
%\wacomment{Can we add ``We release our data splits to help future research in ontology matching.''? If not, other people won't be able to replicate or compare to our results.}
%\lwcomment{Addressed at Introduction: contributions}

\subsection{Deriving definitions and mention contexts}\label{sec:external_data}

Many ontologies do not provide entity definitions (Table \ref{tab:01}). 
In fact, only a few (GO, HPO, MeSH) of the ontologies we included have any definitions at all. 
%Consequently, incorporating a definition embedding into the entity embedding may not be particularly useful. 
%Additionally, if only one of the source or target ontologies includes definitions, the definition embedding may not add any information at all. \wacomment{I don't agree with this last sentence. That would have been the case if our similarity function factorizes across the different attributes, i.e., $G(F_1(\mathbf{v}^s_\textrm{defs}, \mathbf{v}^t_\textrm{defs});F_2(\mathbf{v}^s_\textrm{name},\mathbf{v}^t_\textrm{name}); \ldots)$, but our approximation is more like $G([\mathbf{v}^s_\textrm{defs};\mathbf{v}^s_\textrm{name};\ldots], [\mathbf{v}^t_\textrm{defs};\mathbf{v}^t_\textrm{name};\ldots])$, allowing for interactions between, say, $\mathbf{v}^s_\textrm{defs}$ and $\mathbf{v}^t_\textrm{name}$.} \lwcomment{You're right. We don't forbid interactions between the different parts of the entity embedding, so we would expect partial definitions to contribute as well.}

\begin{table}[ht!]
\begin{center}
\caption{\label{tab:01} Entities with definitions and contexts for each of the training ontologies.}
\vspace{0.8ex}
\begin{tabular}{r|r|r|r}
Ont. & \# of entities & \% w/ def. & \% w/ con. \\
\hline
CPT & 13,786 & 0.0 & 97.9 \\
GO & 44,684 & 100.0 & 30.5 \\
HGNC & 39,816 & 0.0 & 0.8 \\
HPO & 11,939 & 72.5 & 17.9 \\
MeSH & 268,162 & 10.5  & 35.1 \\
OMIM & 98,515 & 0.0 & 2.8 \\
RxNorm & 205,858 & 0.0 & 5.1 \\
\hline
Total & 682,760 & 11.9 & 20.1 
\end{tabular}
\end{center}
\end{table}

We can turn to external sources of entity definitions in such cases. Many biomedical and healthcare concepts are represented in Wikipedia, a general purpose crowd-sourced encyclopedia. The Wikipedia API can be used to search for and extract article content. 
The first paragraph in each Wikipedia article offers a description of the concept, and can be used as a substitute for a definition. 
For each entity in the labeled dataset described in the previous section, we query Wikipedia using the entity's canonical name. 
The first sentence from the top Wikipedia article match is extracted and used to populate the attribute $\mathbf{e}_\textrm{def}$ when undefined in the ontology. For example, a query for OMIM:125370, ``Dentatorubral-pallidoluysian atrophy,'' yields the following summary sentence from Wikipedia: \textit{``Dentatorubral-pallidoluysian atrophy (DRPLA) is an autosomal dominant spinocerebellar degeneration caused by an expansion of a CAG repeat encoding a polyglutamine tract in the atrophin-1 protein.''}
Based on a human-annotated sample, the accuracy of our externally-derived definitions is 75.5\%, based on a random sample of 200 definitions and two annotators with Cohen's kappa coefficient of $\kappa=0.88$.\footnote{\label{annotation}Annotations are available at \url{https://github.com/allenai/ontoemma\#human-annotations}}
%\wacomment{Can we manually check 100 definitions or so to get an estimate of how noisy this was?}

Usage contexts are derived from scientific papers in Medline, leveraging entity annotations available via the Semantic Scholar project \cite{ammar_litgraph_2018}.
In order to obtain the annotations, an entity linking model was used to find mentions of UMLS entities in the abstracts of Medline papers. 
The sentences in which a UMLS entity were mentioned are added to the $\mathbf{e}_{contexts}$ attribute of that entity. 
%For each context in the labeled data, the linked contexts from the Semantic Scholar linking model are extracted for use. 
For UMLS entity C0751781, ``Dentatorubral-Pallidoluysian Atrophy,'' an example context: \textit{``Dentatorubral-pallidoluysian atrophy (DRPLA) is an autosomal dominant neurodegenerative disease clinically characterized by the presence of cerebellar ataxia in combination with variable neurological symptoms,''} is extracted from Yoon et al (2012) \cite{yoon_cerebral_2012}. This context sentence was scored highly by the linking model, and provides additional information about this entity, for example, its acronym (\emph{DRPLA}), the type of disease (\emph{autosomal dominant neurodegenerative}), and some of its symptoms (\emph{cerebellar ataxia}). 
Because there are often numerous linked contexts for each entity, we sample up to 20 contexts per entity when available. The number of entities with context in our labeled data is given in Table \ref{tab:01}.
The accuracy of usage contexts extracted using this approach is 92.5\%, based on human evaluation of 200 contexts with Cohen's kappa coefficient of $\kappa=1$.\footnoteref{annotation}

\section{Experiments}

In this section, we experiment with several variants of OntoEmma:
In the first variant (OntoEmma:NN), we only encode native attributes obtained from the source and target ontologies: canonical name, aliases, and native definitions.
In the second variant (OntoEmma:NN+f), we also add the manually engineered \underline{f}eatures as described in \S\ref{sec:features}.
In the third variant (OntoEmma:NN+f+w), we incorporate external definitions from \underline{W}ikipedia, as discussed in \S\ref{sec:external_data}.
In the fourth variant (OntoEmma:NN+f+w+c), we also encode the usage \underline{c}ontexts we derived from Medline, also discussed in \S\ref{sec:external_data}.
%\wacomment{This is a somewhat serious problem in the experimental design because it doesn't allow us to tell whether the improvements come from external definitions, usage contexts, or engineered features.} \lwcomment{agreed. i believe i ran experiments at each of the different stages: adding engineered features, adding context, adding definitions etc, so i will find and add those results.}

\paragraph{Data.}
We use the training section of the UMLS-derived labeled data to train the model and use the development section to tune the model hyperparameters.
For evaluation, we use the test portion of our UMLS-derived data as well as the OAEI largebio subtrack SNOMED-NCI task, the largest task in OAEI largebio.  The UMLS test set includes 29,859 positive and negative mappings. The OAEI reference alignments included 17,210 equivalent mappings and 1,623 uncertain mappings between the SNOMED and NCI ontologies. 
%\cscomment{How big is the eval set ? i.e. how many test instances? This will help calc the statistical significance one of the reviewers asked for.}
%There are many benchmark datasets for evaluation of ontology matching \cite{ristoski_collection_2016}, but we chose to evaluate the performance of our matcher on biomedical ontologies. \wacomment{omitted because I don't think we need to justify focusing on biomedical ontologies in the bionlp workshop.}

\paragraph{Baselines.}
Our main baseline is a \underline{l}ogistic \underline{r}egression model (OntoEmma:LR) using the same engineered features described in \S\ref{sec:features}.
To illustrate how our proposed method performs compared to previous work on ontology matching, we compare to AgreementMakerLight (AML) which has consistently been a top performer in the OAEI subtasks related to biomedicine \cite{faria_agreementmakerlight_2013}. 
For a fair comparison to OntoEmma, we only use the entity-level matchers in AML; i.e., relation and structural matchers in AML are turned off.\footnote{The performance of the full AML system on the SNOMED-NCI subtask for OAEI 2017 is: precision: 0.90, recall: 0.67, F1: 0.77.} %\cscomment{Do we have a result of the full AML system? It'll probably be valuable to include it at least as a footnote}

\paragraph{Implementation and configuration details.}
We provide an open source, modular, Python implementation of OntoEmma where different candidate selectors, feature generators, and prediction modules can be swapped in and out with ease.\footnoteref{github} 
We implement the neural model using \texttt{PyTorch} and \texttt{AllenNLP}\footnote{\url{https://allennlp.org/}} libraries, and implement the logistic regression model using \texttt{scikit-learn}.
%We use $K=50$ target entities per source entity in the candidate selection step. \wacomment{redundant, already mentioned in model description}
Our 100-dimensional pretrained embeddings are learned using the default settings of \texttt{word2vec} based on the Medline corpus. 
%\cscomment{@Lucy: looks like we added this information here. Just making sure this is correct?}
%\llwcomment{yup! thanks}
%\wacomment{Can we release this?} 
The character-level CNN encoder uses 50 filters of size 4 and 5, and outputs a token embedding of size 100 with dropout probability of 0.2. The LSTMs have output size 100, and have dropout probability of 0.2.
%\wacomment{Add other important hyperparameters used here here.}

\begin{table}[!t]
\caption{Model performance on UMLS test dataset}
\label{tab:umlstest}
\vspace{0.5ex}
\begin{center}
\begin{tabular}{@{}l|c|c|c}
Model & Prec. & Recall & F1 \\
\hline
%AML:entity & & & \\
%\hline
OntoEmma:LR & 0.98 & 0.92 & 0.95 \\
OntoEmma:NN & 0.87 & 0.85 & 0.86 \\
OntoEmma:NN+f & 0.93 & 0.96 & 0.95 \\
OntoEmma:NN+f+w & 0.93 & 0.97 & 0.95 \\
OntoEmma:NN+f+w+c & 0.94 & 0.97 & 0.96 \\
\end{tabular}
\end{center}
\end{table}

\paragraph{Results.}
%\wacomment{Can we report AML results on the UMLS test set as well?}
%\lwcomment{I don't think I have enough time to do this. I would have to edit and rebuild AML. I will move this to the to-do for camera ready.}
The performance of the models is reported in terms of precision, recall and F1 score on the held-out UMLS test set and the OAEI largebio SNOMED-NCI task in Tables \ref{tab:umlstest} and \ref{tab:oaeilargebio}, respectively.

\begin{table}[!t]
\begin{center}
\caption{Model performance on OAEI largebio SNOMED-NCI task}
\label{tab:oaeilargebio}
\vspace{0.5ex}
\begin{tabular}{@{}l|c|c|c}
Model & Prec. & Recall & F1 \\
\hline
AML:entity & 0.81 & 0.62 & 0.70 \\
\hline
OntoEmma:LR & 0.75 & 0.56 & 0.65 \\
%OntoEmma:NN+ & & & \\
%OntoEmma:NN+w & & & \\
OntoEmma:NN+f+w+c & 0.80 & 0.61 & 0.69 \\
\end{tabular}
\end{center}
\end{table}

Table~\ref{tab:umlstest} illustrates how different variants of OntoEmma perform on the held-out UMLS test set.
We note that the bare-bones neural network model (OntoEmma:NN) does not match the performance of the baseline logistic regression model (OntoEmma:LR), suggesting that the representations learned by the neural network are not sufficient. 
Indeed, adding the engineered features to the neural model in (OntoEmma:NN+f) provides substantial improvements, matching the F1 score of the baseline model. 
Adding definitions and usage context in (OntoEmma:NN+f+w+c) further improves the F1 score by one absolute point, outperforming the logistic regression model.

While the UMLS-based test set in Table~\ref{tab:umlstest} represents the realistic scenario of aligning new entities in partially-aligned ontologies, we also wanted to evaluate the performance of our method on the more challenging scenario where no labeled data is available in the source and target ontologies.
This is more challenging because the patterns learned from ontologies used in training may not transfer to the test ontologies.
Table~\ref{tab:oaeilargebio} illustrates how our method performs in this scenario using SNOMED-NCI as test ontologies.
For matching of the SNOMED and NCI ontologies, we enriched the entities first using Wikipedia queries. At test time, we also identified and aligned pairs of entities with exact string matches, using the OntoEmma matcher only for those entities without an exact string match. 
Unsurprisingly, the performance of OntoEmma on unseen ontologies (in Table~\ref{tab:oaeilargebio}) is much lower than its performance on seen ontologies (in Table~\ref{tab:umlstest}).
With unseen ontologies, we gain a large F1 improvement of 4 absolute points by using the fully-featured neural model (OntoEmma:NN+f+w+c) instead of the logistic regression variant (OntoEmma:LR), suggesting that the neural model may transfer better to different domains.
We note, however, that the OntoEmma:NN+f+w+c matcher performs slightly worse than the AML entity matcher. This is to be expected since AML incorporates many matchers which we did not implement in our model, e.g., using background knowledge, acronyms, and other features.
%The addition of usage context did not have discernible effect on model performance, likely due to the small percentage of entity alignments with context, and the distribution of context being primarily with more common entities, which are easy to match based on simple name features alone. todo: report more ablation experiment results here.\wacomment{Visual reminder to add more ablation experiments.}

\section{Discussion}
Through building and training a logistic regression model and several neural network models, we evaluated the possibility of training a supervised machine learning model for ontology alignment based on existing alignment data, and evaluated the efficacy of including definitions and usage context to improve entity matching. 
For the first question, we saw some success with both the logistic regression and neural network models. The logistic regression model performed better than the simple neural network model without engineered features. 
%This outcome came as somewhat of a surprise. Because the LR and NN model used the same entity information, we had anticipated that the increased non-linearity and flexibility of the neural network model would have led to improvements upon hand-engineered features. \wacomment{I strongly disagree with this. Using a NN with raw inputs to outperform simpler models with engineered features typically requires large amounts of labeled data, experimentation with different architectures, ...etc.}
%The better performance of the LR model was likely directly caused by the goodness of our manually-engineered features.
Hand-engineered features encode human knowledge, and are less noisy than features trained from a neural network. The NN model required more training data to learn the same sparse information encoded by pre-defined features. 
%had to handle significantly more dimensions, and  \wacomment{I omitted this because we use sparse features in the LR model which probably means we have the input dimensionality is about the same.}
%Adding the hand-engineered features explicitly to the input of the neural network decision feed forward layer produced comparable results to the logistic regression model.\wacomment{It'd be great to add the results that back up this conclusion to the tables. If not (I know time is limited), please explicitly mention that this conclusion is based on other detailed experiments not reported here.}

To bolster performance, hand-engineered features and extensive querying of third-party resources were used to increase knowledge about each entity. Definitions and usage contexts had rarely been used by previous ontology matchers, and we sought to exploit the value of these additional pieces of information. Definitions especially, often offer information about an entity's relations and attributes, which may not be explicitly defined in the ontology. The ontologies used for training  contained inconsistent information -- some had definitions for all entities, some none; some were well-represented in our context linking model, some were not. To take advantage of such information, therefore, we had to turn to external sources of definitions and contexts, which are understandably more noisy than information provided in the ontology itself.

Using Wikipedia and the Medline corpus, we derived definitions and contexts for many of the entities in the UMLS training corpus. Adding definitions improved the performance of our neural network model. However, high quality definitions native to each terminology would likely have improved results further, since we could not ensure that externally derived definitions were always relevant to the entity of interest.

\paragraph{Limitations.}
Our ontology matcher did not implement any structural matchers, and did not take advantage of relationship data where it existed. In ontologies with well-defined hierarchy or relationships, the structural component provides orthogonal and extremely relevant information for matching. By choosing to focus on entity alignment, we were unable to be competitive on global ontology matching. 

Of all the entities in our UMLS training, development, and test datasets, only 11.9\% of entities had definitions from their source ontology (Table \ref{tab:01}). 
Similarly, we were only able to derive contexts for 20.1\% of the training entities from the Semantic Scholar entity linking model (Table \ref{tab:01}). We were hoping for better coverage of the overall dataset.
We were, however, able to use Wikipedia to increase the overall definition coverage of the entities in our data set to 82.1\%.

Although Wikipedia is a dense resource containing curated articles on many concepts, it is by no means exhaustive. Many of the entities in our training and test data set did not correspond directly to entities in Wikipedia. We also could not review each query to ensure a correct match between the Wikipedia article and the entity. The data is therefore noisy and can introduce error in some cases. Although the overall performance improved upon querying Wikipedia for additional definitions, we believe that dedicated definitions from the source terminologies would perform better where available. 

\paragraph{Future work.} 
%Many ontology matchers use only non-sparse information, such as entity names and the relation hierarchy, for matching. This general methodology fails to account for other pieces of information that may be available to assist in entity alignment, such as entity definitions. Definitions are difficult to use in rule-based alignment systems, and may be sparse (may only be available for one of the two ontologies, or for a small subset of entities). 
We are exploring other ways to derive high-quality definitions from external resources, for example, by deriving definitions from synonymous entities in other ontologies, or by generating textual definitions using the logical definitions given in an ontology. Similarly, we can incorporate usage context from other sources. For example, the Semantic MEDLINE Database (SemMedDB) is a database of semantic relationship predictions from PubMed articles \cite{kilicoglu_semmeddb_2012}. The entity-relation triples in this database can be used to retrieve PubMed article context mapped to UMLS terms.

Continuing on, we aim to develop a more flexible ontology matching system that takes into account all of the information available about an entity. Flexible entity embeddings would represent critical information for proper entity alignment, while accounting for a variety of data types, such as list-like and graph-like data. We would also like to incorporate ontology structure and relations in matching. Hierarchical structure is provided by most biomedical terminologies, and provides essential information for a matching system. One possibility is ensembling OntoEmma with other matcher systems that incorporate or focus on using structural features in matching.
 
%In this work, we focused on the problem of aligning two ontologies for feasibility. However, it should be noted that reducing ontology alignment between N $>$ 2 ontologies to a series of pairwise ontology alignment problems is both suboptimal and nontrivial.
%\lwcomment{not sure we need to include the above; seems kind of out of place.}
 
\paragraph{Related work}
\label{sec:related_work}
The OAEI has been driving ontology matching research in the biomedical domain since 2005. 
It provides evaluation data supporting several tracks such as the anatomy, largebio, and more recently introduced phenotype tracks \cite{faria_oaei_2016}. 
Participating matchers implement a variety of matching techniques including rule-based and statistical methods \cite{faria_oaei_2016,gros_evolution_2016,otero-cerdeira_ontology_2015,shvaiko_ontology_2013}. 
Features used by matchers can be element-level (extracted from each individual entity), or structure-level (based on the topology of the ontology and its relationships). 
Content features can be based on terminology (i.e., names of entities), structure (i.e., how entities are connected), annotations (i.e., annotations made to entities), or reasoning output. 
Some features can also be derived from external sources, such as cross-references to other ontologies, or cross-annotations in other datasets, such as term coincidence in publications, or co-annotation of experiments with terms from different ontologies \cite{husein_review_2016}. 

Notable general purpose matchers that have excelled in biomedical domain matching tasks include AgreementMakerLight (AML), YAM++, and LogMap. 
AML has consistently been a top performer in the OAEI subtasks related to biomedicine. 
It uses a combination of different matchers, such as the lexical matcher (looking for complete string matches between the names of entities), mediating matcher (performing the function of the lexical matcher through a third background ontology), word-based string similarity matcher (matching entities with minimal string edit distances), and others. 
AML then combines these various similarity scores to generate a global alignment between the two input ontologies \cite{faria_agreementmakerlight_2013}. 
YAM++, another successful matcher, implemented a decision tree learning model over many string similarity metrics, but leaves the challenges of finding suitable training data to the user, defaulting to information retrieval-based similarity metrics for its decision-making when no training data is provided \cite{ngo_overview_2016}.
LogMap is a matcher specifically designed to efficiently align large ontologies, generating logical output alignments. The matcher uses high-probability matches as anchors from which to deploy its lexical and structural matchers \cite{jimenez_logmap_2011}.

Our system uses neural networks to learn entity representations and features for matching.
Several published works discuss using neural networks to learn weights over similarity metrics pre-defined by the user or developer of the matching system \cite{djeddi_ontology_2013,peng_ontology_2010,huang_use_2008,hariri_neural-networks-based_2006}. These systems do not use neural networks to generate and learn the features most appropriate for entity matching.
\newcite{qiu_knowledge_2017} proposes and tests an auto-encoder network for unsupervised entity representation learning over a bag of words vector that treats all descriptive elements of each entity (its name, definitions etc.) equally. 
We are interested in investigating how these various descriptive elements contribute to entity matching, how sparsity of specific descriptive fields can be offset by deriving information from external resources, and also whether we can use domain-specific training data to optimize a model for the biomedical domain.

%Although neural networks have been used in the past for the ontology matching task \cite{qiu_knowledge_2017,djeddi_ontology_2013,peng_ontology_2010,huang_use_2008,hariri_neural-networks-based_2006}, it remains unclear how best to learn complex representations for these entities in an end-to-end deep learning model. 

\paragraph{Conclusion}
In this paper, we propose using natural language text associated with entities to improve ontology alignment.
We describe a novel neural architecture for ontology alignment which can encode a variety of information, and derive large amounts of labeled data for training the model.
To address the limited coverage of definitions and usage contexts describing entities, we turn to external resources to supplement the available information about entities in the test ontologies.
Our empirical results illustrate that externally-derived definitions and contexts can effectively be used to improve the performance of ontology matching systems.

\section{Acknowledgements}
We would like to thank the anonymous reviewers for their helpful comments. We also thank John Gennari, Oren Etzioni, Joanna Power as well as the rest of the Semantic Scholar team at the Allen Institute for Artificial Intelligence for helpful comments and insights. 
%Although we have made great headway into the task of ontology and entity alignment, challenges remain. Deep learning has shown successful applications to many domains, we were not able to outperform AML, a top performer in the OAEI competitions. 
%The large difference in performance between seen vs.~unseen ontologies suggest that large improvements can be obtained by adapting the model to the target ontologies.
%This can potentially be done by finding high-confidence alignments in the target ontologies and using them for tuning the model parameters.
%Another challenge is that most ontology matchers, including ours, are unable to incorporate and use all available pieces of information in their matching algorithms. A more flexible system capable of adapting to available information could show improved performance over the status quo.

%\klcomment{is there something wrong w/ the ``Anika Gross et al'' citation?  It says ``Comput Struct Biotechnol J'' when it renders.  Think there are missing characters}
%\lwcomment{That's actually just the abbreviation for the journal. It's correct.}

%\section*{Acknowledgments}

%The authors thank...
%The acknowledgments should go immediately before the references.  Do not number the acknowledgments section ({\em i.e.}, use \verb|\section*| instead of \verb|\section|). Do not include this section when submitting your paper for review.

\bibliography{ontoemma}
\bibliographystyle{acl_natbib}

\end{document}